\documentclass[conference]{IEEEtran}
\IEEEoverridecommandlockouts

\usepackage{microtype}
\usepackage{cite}

\ifCLASSINFOpdf
\usepackage[pdftex]{graphicx}
\DeclareGraphicsExtensions{.pdf,.jpeg,.png,.svg}
\else

\fi

\usepackage{amsmath}
\usepackage{amssymb}
\usepackage{amsfonts}
\usepackage{bm}
\usepackage{mathtools}

\usepackage[ruled,vlined,linesnumbered]{algorithm2e}
\usepackage{algcompatible}
\SetAlCapNameFnt{\small}
\SetAlCapFnt{\small}
\SetAlCapSty{textsc}
\SetAlCapNameSty{textsc}

\usepackage{lettrine}
\usepackage{tikz}
\usetikzlibrary{positioning}

\usepackage{tikz}
\usetikzlibrary{calc,patterns,angles,quotes}
\usetikzlibrary{decorations}
\usetikzlibrary{decorations.pathmorphing,decorations.markings,decorations.pathreplacing}
\usetikzlibrary{positioning, shapes, arrows, calc, backgrounds, angles, quotes, intersections,fit}
\usetikzlibrary{arrows.meta}
\usetikzlibrary{external}

\tikzset{>={Latex[scale=0.65]}, transform shape}
\tikzstyle{very densely dashed} = [dash pattern=on 1.5pt off 1pt]

\newcommand\centerofmass{%
	\tikz[radius=0.2cm] {%
		\fill (0,0) -- ++(0.2cm,0) arc [start angle=0,end angle=90] -- ++(0,-0.4cm) arc [start angle=270, end angle=180];%
		\draw[semithick] (0,0) circle;%
	}%
}

\definecolor{light-gray}{gray}{0.85}

\ifCLASSOPTIONcompsoc
\usepackage[caption=false,font=normalsize,labelfont=sf,textfont=sf]{subfig}
\else
\usepackage[caption=false,font=footnotesize]{subfig}
\fi

\usepackage[acronym]{glossaries}
\newacronym{GNN}{GNN}{Graph Neural Network}
\newacronym{CNN}{CNN}{Convolutional Neural Network}
\newacronym{RNN}{RNN}{Recurrent Neural Network}
\newacronym{MDN}{MDN}{Mixture Density Network}
\newacronym{GMM}{GMM}{Gaussian Mixture Model}
\newacronym{NLP}{NLP}{Natural Language Processing}
\newacronym{CVAE}{CVAE}{conditional variational autoencoder}
\newacronym{GAN}{GAN}{generative adversial network}

\newacronym{GRU}{GRU}{Gated Recurrent Unit}
\newacronym{LSTM}{LSTM}{Long Short-Term Memory}
\newacronym{ELU}{ELU}{Expontential Linear Unit}

\newacronym{IVP}{IVP}{initial value problem}
\newacronym{ODE}{ODE}{Ordinary Differential Equation}
\newacronym{NODE}{neural ODE}{Neural Ordinary Differential Equation}
\newacronym{EKF}{EKF}{Extended Kalman Filter}

\newacronym{IA}{IA}{interaction-aware}
\newacronym{EV}{EV}{ego vehicle}
\newacronym{TV}{TV}{target vehicle}
\newacronym{SV}{SV}{surrounding vehicle}

\newacronym{EWTA}{EWTA}{Evolving Winner Takes All}
\newacronym{NLL}{NLL}{Negative Log-Likelihood}
\newacronym{SWA}{SWA}{Stochastic Weighted Averaging}
\newacronym{RMSE}{RMSE}{Root Mean Squared Error}
\newacronym{ADE}{ADE}{Average Displacement Error}
\newacronym{FDE}{FDE}{Final Displacement Error}
\newacronym{MR}{MR}{Miss Rate}
\newacronym{APDE}{APDE}{Average Path Displacement Error}
\newacronym{ANLL}{ANLL}{Average Negative Log-Likelihood}
\newacronym{FNLL}{FNLL}{Final Negative Log-Likelihood}

\newacronym{MPNN}{MPNN}{Message Passing Neural Network}
\newacronym{GAT}{GAT}{Graph Attention Network}

\newacronym{MDL}{MTP-GO}{Multi-Agent Trajectory Prediction using Graphs and neural ODEs}

\newacronym{CA}{CA}{Constant Acceleration}
\newacronym{CV}{CV}{Constant Velocity}

\newacronym{1XI}{1XI}{Single Integrator}
\newacronym{2XI}{2XI}{Double Integrator}
\newacronym{3XI}{3XI}{Triple Integrator}
\newacronym{CL}{CL}{Curvilinear}
\newacronym{ST}{ST}{Kinematic Single-Track}
\newacronym{U}{UC}{Unicycle}
\newacronym{C}{CT}{Curvature}

\newacronym{NODE1}{N-ODE$1$}{First Order}
\newacronym{NODE2}{N-ODE$2$}{Second Order}

\newacronym{EF}{EF}{forward-Euler method}
\newacronym{MP}{MP}{Midpoint method}
\newacronym{Heun}{Heun}{Heun's method}
\newacronym{RK3}{RK3}{Kutta's third-order method}
\newacronym{SSRPK3}{SSRPK3}{third-order strong stability preserving Runge-Kutta method}
\newacronym{RK4}{RK4}{classic fourth-order Runge-Kutta method}
\newacronym{RK45}{DOPRI}{Dormand-Prince variable-step method}
\newacronym{IMA}{Adams}{implicit-Adams method}

\newcommand{\highd}{\emph{highD}}

\newcommand{\round}{\emph{rounD}}

\newcommand{\tplusplus}{Trajectron++}
\newcommand{\mdl}{MTP-GO}

\newcommand{\graph}{\mathcal{G}}
\newcommand{\node}{\mathcal{V}}
\newcommand{\edge}{\mathcal{E}}
\newcommand{\agent}{\nu}

\newcommand{\feature}[2][\agent{}]{\bm{f}_{#2}^{#1}}

\newcommand{\pos}[2][\nu]{\bm{x}^{#1}_{#2}}
\newcommand{\predpos}[2][\nu]{\hat{\bm{x}}^{#1}_{#2}}
\newcommand{\covest}[2][\nu]{\bm{P}^{#1}_{#2}}

\newcommand{\stcov}[1]{\bm{P}_{#1}}
\newcommand{\stjac}[1]{\bm{F}_{#1}}
\newcommand{\inpjac}[1]{\bm{G}_{#1}}

\newcommand{\pnoisem}[1]{\bm{Q}_{#1}}
\newcommand{\dt}{T_s}

\newcommand{\h}{h}
\newcommand{\slope}[1]{\eta_{#1}}

\newcommand{\mix}[1]{\bm{\pi}^{#1}}

\newcommand{\inp}[1]{\bm{u}_{#1}}
\newcommand{\state}[1]{\bm{x}_{#1}}
\newcommand{\stateestim}[1]{\hat{\bm{x}}_{#1}}

\newcommand{\std}{\sigma}
\newcommand{\corr}{\rho}

\newcommand{\history}{\mathcal{H}}
\newcommand{\predtime}{t}
\newcommand{\predhistory}{t_h}
\newcommand{\predhrz}{t_f}

\newcommand{\hidden}{\bm{h}}
\newcommand{\encoding}{\bm{o}}
\newcommand{\hiddendim}{d_h}
\newcommand{\featuredim}{d_f}

\newcommand{\encrep}[2][\agent]{\hidden_{#2}^{#1}}
\newcommand{\encfull}[2][\agent]{\encoding_{#2}^{#1}}
\newcommand{\grurepx}[2][\agent]{\bm{\kappa}_{#2,i}^{#1}}
\newcommand{\grureph}[2][\agent]{\bm{\xi}_{#2,i}^{#1}}

\newcommand{\gnnf}[2][\agent{}]{\text{GNN}_{f}\left(\feature[#1]{#2}, \set{\feature[\tau]{#1}}_{\tau \neq{} #1} \right)}
\newcommand{\gnnh}[2][\agent{}]{\text{GNN}_{h}\left(\encrep[#1]{#2}, \set{\encrep[\tau]{#2}}_{\tau \neq #1} \right)}

\DeclareMathOperator*{\argmin}{argmin}
\DeclareMathOperator*{\argmax}{argmax}

\newcommand{\R}{\mathbb{R}}
\newcommand{\set}[1]{\left\{#1\right\}}

\newcommand{\tuple}[1]{\left(#1\right)}
\newcommand\norm[1]{\lVert#1\rVert}

\newcommand{\transpose}{\text{T}} %

\newcommand{\eq}[2][my_equation]{\begin{equation}\label{eq:#1}#2\end{equation}}
\newcommand{\al}[2][my_equation]{\begin{align}\label{eq:#1}#2\end{align}}

\usepackage{float}
\usepackage{xargs}
\usepackage[colorinlistoftodos,prependcaption,textsize=tiny]{todonotes}
\newcommandx{\unsure}[2][1=]{\todo[linecolor=red,backgroundcolor=red!25,bordercolor=red,#1]{#2}}
\newcommandx{\change}[2][1=]{\todo[linecolor=blue,backgroundcolor=blue!25,bordercolor=blue,#1]{#2}}
\newcommandx{\info}[2][1=]{\todo[linecolor=OliveGreen,backgroundcolor=OliveGreen!25,bordercolor=OliveGreen,#1]{#2}}
\newcommandx{\improvement}[2][1=]{\todo[linecolor=blue,backgroundcolor=blue!25,bordercolor=blue,#1]{#2}}

\usepackage[hidelinks]{hyperref}
\usepackage{cleveref}
\Crefname{figure}{Fig.}{Figs}
\crefname{figure}{Fig.}{Figs}
\crefname{algorithm}{Algorithm}{Algorithms}
\crefname{table}{Table}{Tables}
\crefname{section}{Section}{Sections}
\crefname{equation}{}{}

\usepackage{multirow}
\usepackage{booktabs}
\usepackage{tabu}

\usepackage{url}

\begin{document}

		\title{Evaluation of Differentially Constrained Motion Models for Graph-Based Trajectory Prediction}
		
		\author{Theodor Westny\IEEEauthorrefmark{1}, Joel Oskarsson\IEEEauthorrefmark{2}, Bj\"orn Olofsson\IEEEauthorrefmark{3}\IEEEauthorrefmark{1}, and Erik Frisk\IEEEauthorrefmark{1} 					
		\thanks{\hspace{-0.1in}This research was supported by the Strategic Research Area at Linköping-Lund in Information Technology (ELLIIT),
			the Swedish Research Council via the project \emph{Handling Uncertainty in Machine Learning Systems} (contract number: 2020-04122),
			and the Wallenberg AI, Autonomous Systems and Software Program (WASP) funded by the Knut and Alice Wallenberg Foundation.}
		\thanks{\scriptsize\IEEEauthorrefmark{1}Department of Electrical Engineering,
			Linköping University, Sweden.}
		\thanks{\scriptsize\IEEEauthorrefmark{2}Department of Computer and Information Science,
			Linköping University, Sweden.}
		\thanks{\scriptsize\IEEEauthorrefmark{3}Department of Automatic Control,
			Lund University, Sweden.}
		}
		\maketitle

		\begin{abstract}
			Given their flexibility and encouraging performance, deep-learning models are becoming standard for motion prediction in autonomous driving.
			However, with great flexibility comes a lack of interpretability and possible violations of physical constraints.
			Accompanying these data-driven methods with differentially-constrained motion models to provide physically feasible trajectories is a promising future direction.
			The foundation for this work is a previously introduced graph-neural-network-based model, MTP-GO.
			The neural network learns to compute the inputs to an underlying motion model to provide physically feasible trajectories.
			This research investigates the performance of various motion models in combination with numerical solvers for the prediction task.
			The study shows that simpler models, such as low-order integrator models, are preferred over more complex, e.g., kinematic models, to achieve accurate predictions.
			Further, the numerical solver can have a substantial impact on performance, advising against commonly used first-order methods like Euler forward.
			Instead, a second-order method like Heun's can greatly improve predictions.
		\end{abstract}

		\IEEEpeerreviewmaketitle

		\section{Introduction}
The unknown decisions of surrounding road users are a primary source of uncertainty in traffic situations.
Similarly to how a human driver adapts its future trajectory based on anticipations of the environment---autonomous vehicles should be equipped with the ability to predict the future actions of other traffic participants in order to ensure safe and proactive operation.
The \emph{behavior prediction} task \cite{mozaffari2020deep} encapsulates this problem of predicting the intention and future motion of surrounding traffic agents, and its importance as a research topic has grown significantly over the last decades.

Due to the considerable difficulty of hand-crafting models that can decode the social interactions between a time-varying number of traffic participants, learning-based approaches have offered useful adaptability, proving valuable in addressing these complex problems \cite{huang2022survey}. 
Despite their flexibility, however, learning-based methods exhibit certain limitations. 
First, unlike conventional state estimation, these methods often lack interpretability because of the numerous latent representations. 
More importantly, they rarely provide performance guarantees, rendering them less attractive in applications with stringent safety requirements.
Recent research has proposed coupling (deep) data-driven models with differential constraints to address these issues within the scope of motion prediction.
The general idea is to have the learnable components of the model compute the inputs to an underlying motion model, thereby generating physically feasible outputs \cite{cui2020deep, salzmann2020trajectron, li2021spatio, westny2023graph}.
By drawing inspiration from target tracking \cite{li2003survey}, or model-based control \cite{paden2016survey}, numerous formulations can be utilized for the motion prediction task, including various orders of integrators, non-holonomic constrained models or even \glspl{NODE} \cite{chen2018neuralode}.

Leveraging on our work in \cite{westny2023graph}, this paper delves deeper into integrating deep graph-based networks with differentially-constrained motion models for trajectory prediction.
The investigation focuses on the design of differential constraints, appropriate complexity class selection for heterogeneous traffic scenarios, and their impact on prediction and training performance.
Furthermore, as numerical solvers are used to compute model states, the choice of integration methods can significantly influence the outcomes. 
In light of this, an investigation into the consequences of choosing various numerical solvers in combination with different motion models is conducted.

\begin{figure}[!t]
	\begin{tikzpicture}[>=latex, transform shape]
    \tikzset{wheel/.style = {draw, rounded corners=1mm, minimum height=0.4cm,    minimum width=1.2cm, ultra thick}}
    \draw[->,thick] (0,0) coordinate (O) --(7.5,0) coordinate (X) node[below]{$\mathbf X$};
    \draw[->,thick] (0,0)--(0,3.2) coordinate (Y) node[left]{$\mathbf Y$};

    \begin{scope}[rotate around={15:(1,1)}]
      \node [wheel] (wr) at (1,1) {};
      \node [wheel, rotate=40, right=5cm of wr, anchor=center] (wf) {};

      \draw [thick] (wr) -- (wf) node [pos=0.4] (com) {\centerofmass};
      \draw [->, very thick] (com.center) -- ++(25:1.5cm) coordinate (v) node [near end, above=1mm] {$v$};
      \draw pic["$\beta$",<->,angle eccentricity=1.25,angle radius=1.2cm,draw] {angle=wf--com--v};

      \draw [dashed] (wf.center) -- +(40:-1cm) coordinate (deltaf1);

      \draw pic["$u_1$",<->,angle eccentricity=1.3,angle radius=0.8cm,draw] {angle=wr--wf--deltaf1};

      \coordinate [below=5mm of wr] (wrbelow);
      \draw [densely dashed] (wr) -- (wrbelow);
      \draw [densely dashed] (com.south) -- (wrbelow-|com.south);
      \draw [densely dashed] (wf) -- (wrbelow-|wf);

      \draw [<->] (wrbelow-|wf) -- node [below, midway] {$l_f$} (com|-wrbelow);
      \draw [<->] (wrbelow) -- node [below] {$l_r$} (com|-wrbelow);
    \end{scope}

    \draw[dotted] (Y|-com) node [left] {$Y$} -- (com.center) -- ++(1.5cm,0) coordinate (psi);
    \draw[dotted] (X-|com) node [below] {$X$} -- (com.center);

    \draw pic["$\psi$",<->,angle eccentricity=1.3,angle radius=1.2cm,draw] {angle=psi--com--wf};

  \end{tikzpicture}
	\vspace{-0.05in}
	\caption{Schematics of the kinematic single-track model with $u_1$ denoting the steering angle.
	 Its use for motion prediction is interesting due to its connection and frequent use in autonomous planning and control applications\cite{paden2016survey}.}
	\label{fig:mdl}
\end{figure}
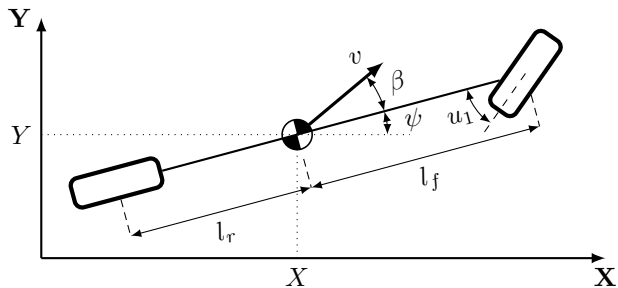
\subsection{Contributions} 
The primary contributions of this paper are:
\begin{itemize}
	\item An investigation into the properties of differentially constrained motion models across distinct complexity classes, illustrating their impact on overall model effectiveness.
	\item A study of the effects of diverse continuous-time integration techniques used in dynamic motion models, and their influence on training efficacy and prediction accuracy.
\end{itemize}
All investigations were performed based on the \gls{GNN} architecture \mdl~\cite{westny2023graph} for prediction performance using the \highd{}~\cite{highDdataset} and \round{}~\cite{rounDdataset} data sets.
Implementations are available online\footnote{ \url{https://github.com/westny/mtp-go}}.
		\section{Related Work}
\label{sec:related_work}
The motion prediction problem can be found in many research fields, closely related to applications in state estimation and target tracking \cite{huang2022survey}.
Predicting the future states of the ego-vehicle is useful for the development of active safety systems.
In such applications, physics-based models are typically employed to obtain physically feasible predictions.
In \cite{lin2000vehicle}, a vehicle-dynamics model in combination with a Kalman filter is used to predict future position.
In \cite{polychronopoulos2007sensor}, the use of different physics-based vehicle models is presented for collision avoidance applications.
An investigation on a range of curvilinear motion models \cite{li2003survey} for vehicle tracking is presented in \cite{schubert2008comparison}, illustrating the importance of appropriate model selection dependent on the application.
While physics-based methods generalize well, they often simplify the true dynamics.
One possibility entails using a grey-box approach to combine the physical model with learnable parts, for example, using Gaussian processes \cite{kullberg2021online}.
Combined with Kalman filtering, physics-based models are suitable for applications with short-term predictions. 
However, when the prediction horizon increases, assumptions on motion-model inputs become increasingly invalid, requiring time-varying input predictions conditioned on the current scene.

Recently, the motion prediction problem has been the target of learning-based research \cite{mozaffari2020deep, huang2022survey}.
Because of the temporal nature of the problem, several approaches have based their models on \glspl{RNN}~\cite{hu2018probabilistic, messaoud2020attention, deo2018convolutional, li2019grip, salzmann2020trajectron, li2021spatio} or more recently, Transformers~\cite{liu2021multimodal, huang2022multi}.
To utilize the spatial characteristics of the problem, these models are often integrated with \glspl{CNN} \cite{deo2018convolutional, li2019grip}, or \glspl{GNN} \cite{li2019grip, jeon2020scale, salzmann2020trajectron, li2021spatio, hu2022scenario}.
A potential issue with complete black-box models aimed at predicting future motion is that model outputs can be physically infeasible.
In \cite{cui2020deep}, the use of a kinematic single-track model \cite{paden2016survey} is proposed for inclusion in the prediction model.
The idea is to have a deep neural network compute the inputs to the motion model to generate kinematically feasible outputs.
\tplusplus \cite{salzmann2020trajectron} is a \gls{GNN}-based method that does trajectory prediction by a recurrent generative model combined with model-based kinematic constraints.
In the paper, a modified unicycle model is used to describe wheeled vehicles and a single-order integrator is used to describe pedestrians.
STG-DAT \cite{li2021spatio} is a similarly structured model to that of \tplusplus{}. %
Similarly to \cite{cui2020deep}, STG-DAT employs a kinematic single-track model in the prediction model.
In \cite{westny2023graph}, \mdl{} is proposed.
The method uses an encoder--decoder model based on temporal \glspl{GNN} to compute the motion model inputs.
Instead of using predetermined motion constraints, the motion models are learned using \glspl{NODE}.

		\section{Problem formulation}
\label{sec:problem_formulation}
The trajectory prediction problem is formulated as estimating the probability distribution of the future positions of all agents $\agent \in \node_{\predtime}$ currently in the scene for each time instant $\predtime+1, \dots, \predtime+\predhrz$.
The predicted mean future trajectory of an agent $\agent$ is a sequence $\predpos{\predtime+1}, \dots, \predpos{\predtime+\predhrz}$ of time-stamped positions in $\R^2$.
The forecasted trajectory is accompanied by an estimated state covariance  $ \covest{\predtime+1}, \dots, \covest{\predtime+\predhrz}$ used to represent the prediction uncertainty.
To provide dynamically feasible outputs, the future trajectories are computed using differentially constrained motion models $f$ where
\begin{equation}
	\label{eq:mmodel}
	\dot{\state{}} = f(\state{}, \inp{}),
\end{equation}
and the input $\inp{}$ is the output of a deep neural network.

The two main research questions of this paper concern the formulation and integration of the motion model in the trajectory predictor. 
Possible model assumptions in \cref{eq:mmodel} range from single integrators to kinematic models and neural ODEs.
How this choice affects performance and training is a central research question.
Since the models are formulated using differential equations, model states are retrieved using numerical integration methods.
Therefore, the impact of the choice of numerical \gls{ODE} solver on training and prediction performance is researched.
All investigations are performed based on a \gls{GNN} architecture \cite{westny2023graph} and are evaluated using naturalistic driving data \cite{highDdataset,rounDdataset}.

		\section{Graph-based Traffic Modeling}
Based on the proposal in \cite{westny2023graph}, a traffic situation over $n$ time steps is modeled as a sequence $\graph_1, \dots, \graph_n$ of graphs centered around a vehicle $\agent_0$.
For a graph $\graph_i = (\node_i, \edge_i)$, the sets $\node_i$ and $\edge_i$ refer to the agents currently in the scene and their edges, respectively.
Given an agent $\agent \in \node_i$, the model has access to its historic observations $\feature{i} \in \R^{\featuredim}$, such as previous planar positions and velocities (see \cref{tab:node_feat}) from time $\predtime - \predhistory$ until $\predtime$.
The graphs within the observation window may be dissimilar at different time instants because of the arrival and departure of agents.
Predictions are computed for all the agents $\node_{\predtime}$ still in the scene at prediction time $\predtime$.
Given the full history
\eq[traffic_history]{
	\history = \left(
	\set{\graph_i}_{i=\predtime - \predhistory}^{\predtime},
	\set{\set{\feature{i}}_{\agent \in \node_i}}_{i=\predtime - \predhistory}^{\predtime}
	\right)
}
probabilistic trajectory prediction can then be summarized as modeling the conditional distribution
\eq[traj_pred_dist]{
	p\left(\set{\tuple{\pos{\predtime+1}, \dots, \pos{\predtime+\predhrz}}}_{\agent \in \node_{\predtime}} \middle| \history \right).
}
While the feature histories of different nodes can be of different lengths at the prediction time instant $\predtime$, the model should still be able to predict the future trajectory for all nodes currently in the graph.

\begin{table}[!t]
	\vspace{-0.1in}
	\caption{Input features}
	\label{tab:node_feat}
	\centering
	\begin{tabular}{c l l}
		\toprule
		Feature & Description & Unit\\
		\midrule
		$x$ & Longitudinal coordinate & m\\
		$y$ & Lateral coordinate & m\\
		$v_x$ & Instantaneous longitudinal velocity & m/s \\
		$v_y$ & Instantaneous lateral velocity & m/s\\
		$a_x$ & Instantaneous longitudinal acceleration & m/$\text{s}^2$ \\
		$a_y$ & Instantaneous lateral acceleration & m/$\text{s}^2$\\
		$\psi$ & Yaw angle  & rad \\
		\midrule
		\multicolumn{3}{c}{\emph{Highway specific} \cite{westny2021vehicle}} \vspace{0.05in}\\
		$d_y$ & Lateral deviation from the current lane centerline & $[-1, 1]$\\
		$d_r$ & Lateral deviation from the road center & $[-1, 1]$\\
		\midrule
		\multicolumn{3}{c}{\emph{Roundabout specific} \cite{westny2023graph}} \vspace{0.05in}\\
		$r$ & Euclidean distance from the roundabout center & m\\
		$\theta$ & Angle relative to the roundabout center & rad\\
		\bottomrule
	\end{tabular}
\end{table}

		\section{Motion Modeling}
\label{sec:motion-modeling}
The prediction model consists of a deep neural network that learns to predict the inputs $\inp{} = [u_1, u_2]$ of a differentially constrained motion model with states $\state{}$.
Several motion models used in target tracking \cite{li2003survey} and predictive control \cite{paden2016survey} applications hold potential for use in behavior prediction.
This research aims to investigate alternative formulations and draw conclusions regarding their usability and effectiveness.
The models considered are of different orders, but they all have at least two state variables $[x, y]$, used to represent the planar positions of the prediction target.

\subsection{Pure Integrators}
Pure-integrator models are the simplest considered here.
The inputs enter into the differential equations with the highest derivative, which depends on the model degree.
Intermediate state transitions are modeled as direct integrations.

\subsubsection{\gls{1XI}}
The model has two states ($x, y$), and their rate of change can be directly controlled by the neural network, i.e., velocities $v_x, v_y$ are control signals $u_1, u_2$, respectively as
\begin{subequations}
\begin{align}
	\dot{x} &= u_1 \\
	\dot{y} &= u_2
\end{align}
\end{subequations}

\subsubsection{\gls{2XI}}
The model has four states, positions and velocities.
The accelerations can be controlled by the network:
\begin{subequations}
\begin{align}
		\dot{v}_x &= u_1 \\
		\dot{v}_y &= u_2
\end{align}
\end{subequations}

\subsubsection{\gls{3XI}}
The model has the most states (six) of all considered. 
In this case, the neural network controls the jerk: %
\begin{subequations}
\begin{align}
		\dot{a}_x &= u_1 \\
		\dot{a}_y &= u_2
\end{align}
\end{subequations}

\subsection{Orientation-Based Models}
Orientation-based models refer to motion models with internal states of orientation $\psi$ and speed $v$, generally formulated as
\begin{subequations}
	\label{eq:modeling-orientation}
\begin{align}
	\dot{x} &= v \cos(\psi) \\
	\dot{y} &= v \sin(\psi) \\
	\dot{\psi} &= \chi(\bm{x}, \bm{u}) \\
	\dot{v} &= u_2,
\end{align}
\end{subequations}
where the driving function $\chi$ varies by chosen model, stated explicitly in the respective descriptions.

\subsubsection{\gls{CL}}
This refers to a general, curvilinear-motion model \cite{li2003survey}:
\begin{align}
	\dot{\psi} &= \frac{u_1}{v},
\end{align}
where the input $u_1$ represents the acceleration perpendicular to the trajectory.

\subsubsection{\gls{C}}
Mathematically, the curvature formulation looks similar to the curvilinear-motion model.
However, the input does not refer to the acceleration but instead the curvature of the current trajectory arc:
\begin{align}
	\dot{\psi} &= u_1 v 
\end{align}

\subsubsection{\gls{U}}
The unicycle model represents the vehicle as a single controllable wheel, where the change in turn rate is one of the inputs:
\begin{align}
	\dot{\psi} &= u_1
\end{align}
In \cite{salzmann2020trajectron}, this is used to describe wheeled vehicles.

\subsubsection{\gls{ST}}
The kinematic single-track model, illustrated in \cref{fig:mdl}, is commonly used in motion planning and control applications \cite{paden2016survey}.
Here, the orientation is controlled by the steering angle $u_1$:
\begin{subequations}
\begin{align}
	\dot{x} &= v \cos(\psi + \beta) \\
	\dot{y} &= v \sin(\psi + \beta) \\
	\dot{\psi} &= \frac{v}{l_r} \sin(\beta) \\
	\beta &= \arctan\left(\frac{l_r}{l_f + l_r}\tan(u_1)\right)
\end{align}
\end{subequations}
In this research, the lengths $l_f$ and $l_r$ that make up the wheelbase are estimated using the current vehicle's dimensions.
Note that the angle $\beta$ enters into the equations describing $\dot{x}$ and $\dot{y}$, which is different from \cref{eq:modeling-orientation}.
A slightly modified version of this model is used in \cite{li2021spatio}.

\subsection{Neural \acrlong{ODE}s}
A \gls{NODE} is a type of neural network that learns a continuous-time dynamic system by modeling its derivatives \cite{chen2018neuralode}.
Adopting the methods proposed in our previous work \cite{westny2023graph}, two general \gls{NODE} formulations of varying order are considered.
The motion models consist of differentially-constrained feedforward networks with learnable parameters.

\subsubsection{\gls{NODE1}}
Two separate \glspl{ODE}, $f_1$ and $f_2$ are used to describe the state dynamics where each function is associated with its respective input:
\begin{subequations}
\begin{align}
		\dot{x} &= f_1(x, y, u_1) \\
		\dot{y} &= f_2(x, y, u_2)
\end{align}
\end{subequations}

\subsubsection{\gls{NODE2}}
For the second-order model, only the highest-order derivatives are parameterized. 
Intermediate states are pure integrators, which gives
\begin{subequations}
\begin{align}
		\dot{x} &= v_x \\
		\dot{y} &= v_y \\
		\dot{v}_x &= f_1(v_x, v_y, u_1) \\
		\dot{v}_y &= f_2(v_x, v_y, u_2)
\end{align}
\end{subequations}

\subsection{Generating Dynamically Feasible Model Inputs}
Although the neural network may directly learn which ranges of model inputs are suitable for the best performance, outputting feasible values can not be guaranteed unless explicitly handled.
Additionally, bounding the inputs further supports the goal of guaranteeing feasible outputs. 
Bounds on the motion-model inputs are enforced using the HardTanh activation function:
\begin{equation}
	\text{HardTanh}_{u}(u) = \begin{cases}
		u_{\rm max}, \quad &u > u_{\rm max} \\
		u_{\rm min},\quad &u < u_{\rm min} \\
		u, \quad &\text{otherwise}
	\end{cases}
\end{equation}
Most input bounds are based on the training data, if available; otherwise, they are determined by physical insight.
For simplicity, the double-sided constraint is assumed to be symmetric, i.e., $u_{\rm min} = -u_{\rm max}$, which is reasonable for all inputs.

\subsection{Numerical Integration}
The motion models presented in \cref{sec:motion-modeling} are all formulated as controllable, continuous-time \glspl{ODE}.
The model states are retrieved by solving an initial-value problem using methods of numerical integration.
In this research, how the choice of these methods affects the prediction performance is investigated, most of which are in the Runge-Kutta family of methods \cite{ascher1998computer}.
Consider the general model in \cref{eq:mmodel}, stated again for convenience:
\begin{equation}
	\dot{\state{}} = f(\state{}, \inp{})
\end{equation}
The perhaps most well-known \gls{ODE} solver is the forward-Euler method, which for a given step size $\h$ is formulated:
\begin{equation}
	\label{eq:euler}
	\state{k+1} = \state{k} + h f(\state{k}, \inp{k})
\end{equation}
Due to its simplicity and low computational cost, the forward-Euler method is an appealing choice for solving differential equations.
It is also the choice of method in \cite{li2021spatio}.
However, the forward-Euler method does have drawbacks, most notably a small region of stability, heavily dependent on the appropriate choice of step size $\h$ \cite{ascher1998computer}.
Other methods, specifically those of higher order, are typically favored in practical applications where accuracy is more important.
The classic fourth-order Runge-Kutta method is one such example \cite{ascher1998computer}:
\begin{subequations}
	\label{eq:rk4}
	\begin{align}
		\slope{1} &= f(\state{k}, \inp{k}), \\
		\slope{2} &= f(\state{k} + \frac{\h}{2} \slope{1}, \inp{k}), \\
		\slope{3} &= f(\state{k} + \frac{\h}{2} \slope{2}, \inp{k}), \\
		\slope{4} &= f(\state{k} + \h \slope{3}, \inp{k}), \\
		\state{k+1} &= \state{k} + \frac{\h}{6} (\slope{1} + 2 \slope{2} + 2 \slope{3} + \slope{4})
	\end{align}
\end{subequations}
In numerical analysis software, it is generally not a fixed-step method like those stated in this subsection that is the default. 
Instead, it is typically a variable-step solver, such as the Dormand-Prince method \cite{dormand1980family}.
In these solvers, the step size $\h$ is computed based on intermediate calculations.
		\section{Trajectory Prediction Model}
\label{sec:model}
This work leverages the \mdl{}\footnote{\textbf{M}ulti-agent \textbf{T}rajectory \textbf{P}rediction by \textbf{G}raph-enhanced neural \textbf{O}DEs} model presented in our previous research. %
Therefore, only a brief summary of the main components will be presented here; for details, see \cite{westny2023graph}.
The complete \mdl{} model consists of a \gls{GNN}-based encoder--decoder module that computes the inputs to a motion model for trajectory forecasting.
The output is multi-modal, consisting of several candidate trajectories $\stateestim{\predtime+1}^j, \dots, \stateestim{\predtime + \predhrz}^j$ for different components $j\in\{1, \dots, M\}$ used to capture different maneuvers.
In addition, each candidate is accompanied by a predicted state covariance $\stcov{\predtime+1}^j, \dots, \stcov{\predtime + \predhrz}^j$ which is estimated using an \gls{EKF}.

\subsection{Temporal \acrlong{GNN} Encoder}
\subsubsection{Graph-Gated Recurrent Unit}
Using an extended \gls{GRU} cell \cite{cho2014properties} where the conventional linear mappings are replaced by \gls{GNN} components~\cite{oskarsson2022temporal}, spatial-temporal interactions can be captured. %
The \glspl{GNN} takes as input the representations for the specific node $\agent$ and the information of other nodes in the graph.
Intermediate representations are computed by two \glspl{GNN} as
\begin{subequations}
\label{eq:gnn_interm}
\al[gnn_interm_f]{
    \left[\grurepx{r} \| \grurepx{z} \| \grurepx{h} \right] &=
        \gnnf{i}\\
    \label{eq:gnn_interm_h}
    \left[\grureph{r} \| \grureph{z} \| \grureph{h} \right] &=
        \gnnh{i-1},
}
\end{subequations}
where $\|$ is the concatenation operation.
These are then used to compute the representation $\encrep{i}$ for time step $i$ as
\begin{subequations}
\begin{align}
	\label{eq:full_gru_update}
	\bm{r}_i^\agent &= \sigma(\grurepx{r} + \grureph{r} + \bm{b}_r) \\
	\bm{z}_i^\agent &= \sigma(\grurepx{z} + \grureph{z} + \bm{b}_z) \\
	\bm{\tilde{h}}_i^\agent &= \phi(\grurepx{h} + \bm{r}_i^\agent \odot \grureph{h} + \bm{b}_{h})\\
	\encrep{i} &= (\bm{1} - \bm{z}_i^\agent) \odot \bm{\tilde{h}}_i^\agent{} + \bm{z}_i^\agent \odot \encrep{i-1},
\end{align}
\end{subequations}
where the bias terms $\bm{b}_r$, $\bm{b}_z$, $\bm{b}_h \in \R^{\hiddendim}$ are additional learnable parameters, $\odot$ the Hadamard product, $\sigma$ the sigmoid function, and $\phi$ is the hyperbolic tangent.

\subsubsection{\acrlong{GNN}}
In \cite{westny2023graph}, the properties of different \gls{GNN} layers and their use in trajectory prediction are investigated.
Here, the \glspl{GNN} in \cref{eq:gnn_interm} are modeled using a modified version of the \gls{GAT} \cite{brody2022attentative} framework.
To compute the aggregation of weights for a neighborhood, \gls{GAT} layers utilize an attention mechanism \cite{vaswani2017attention}.
In our modified version, called \emph{\gls{GAT}+}, the results of the standard \gls{GAT} output are summed with the output of a linear layer that takes the representation of the center node \cite{westny2023graph}.

\subsection{Decoder with Temporal Attention Mechanism}
Similarly to the encoder, the decoder also utilizes a graph-\gls{GRU} component.
Its task is to compute the motion model input $\inp{}$ and the process noise $\pnoisem{}$ for time steps $\predtime+1, \dots, \predtime + \predhrz$.
The first $\bm{h}$-input to the decoder is taken as the last hidden representation from the encoder $\encrep{\predtime}$.
The updates then proceed just as in \cref{eq:gnn_interm_h}.
To construct the input $\bm{f}$, the decoder additionally utilizes a temporal attention mechanism conditioned on the full encoder representation $\encfull{\predtime}=[\encrep{\predtime - \predhistory}, \dots \encrep{\predtime}]$.

\subsection{Uncertainty Propagation in Forecasting}
The prediction step of the \gls{EKF} is used to estimate the future states and state covariances.
Combining the computed estimates with a mixture density network\cite{bishop2006pattern}, the model outputs multi-modal future predictions by learning the parameters of a Gaussian mixture model\cite{bishop2006pattern}.
\subsubsection{Extended Kalman Filter}
For a differentiable state-transition function $f$, the prediction step of the \gls{EKF} is:
\begin{subequations}
	\begin{align}
		\stateestim{k+1} &= f(\stateestim{k|k}, \inp{k}) \\
		\stcov{k+1} &= \stjac{k} \stcov{k|k} \stjac{k}^\transpose + \inpjac{k} \pnoisem{k|k} \inpjac{k}^\transpose,
	\end{align}
\end{subequations}
where
\begin{equation}
	\stjac{k} = \frac{\partial f}{\partial \state{}}\Bigr|_{\stateestim{k|k}, \inp{k}}
\end{equation}
Here, $\hat{\bm{x}}$ and $\bm{P}$ refer to the state estimate and state covariance estimate, respectively, and $f$ is the current motion model.
For all motion models, the process noise is assumed to be a consequence of the predicted inputs and therefore enters into the two highest-order states.
This noise is modeled as zero-mean with covariance matrix
\begin{equation}
	\bm{Q} = \begin{pmatrix}
		\std_{1}^2 & \corr \std_{1} \std_{2} \\
		\corr \std_{1} \std_{2} & \std_{2}^2
	\end{pmatrix}
\end{equation}
Assuming that the noise is additive, $\inpjac{k}$ is designed as a matrix of constants.
In the simplest case with two state variables, then $\inpjac{k} = \dt \cdot I_{2}$, where $\dt$ is the sample period.
For higher-order state-space models, $\inpjac{k}$ is generalized:
\begin{equation}
	\inpjac{k}=  \dt
	\begin{pmatrix}
		 0 & \cdots & 1 & 0 \\
		 0 & \cdots & 0 & 1
	\end{pmatrix}^\transpose{}
\end{equation}

\subsubsection{\acrlong{MDN}}
Each output vector $\bm{y}_k$ of the model contains mixing coefficients $\mix{j}$, along with the state estimate $\stateestim{k}^j$ and state covariance estimate $\stcov{k}^j$ for all mixtures $j\in\{1, \dots, M\}$:
\begin{equation}
	\label{eq:gmm}
	\bm{y}_k = \left( \mix{j}, \big\{\stateestim{k}^j,  \stcov{k}^j\big\}_{j=1}^{M} \right),
\end{equation}
where $\mix{j}$ is constant over the prediction horizon $\predhrz$.
For notational convenience, the predictions are indexed from $k=1, \dots, \predhrz$.
The model is trained by minimizing the \gls{NLL} of the ground truth trajectory
\begin{equation}
	\label{eq:seq_loss}
	\mathcal{L}_{\text{NLL}} =\sum_{k= 1}^{\predhrz} -\log \left(\sum_j \mix{j}\mathcal{N}(\state{k} | \stateestim{k}^j, \stcov{k}^j)\right)
\end{equation}

		\section{Evaluation \& Results}
\label{sec:results}
The model combinations are evaluated using different data sets, encouraged by their varying dynamics and behavior.
To motivate the use of a deep-learning-based backbone to compute the motion model inputs, a \gls{CA} and \gls{CV} model is included as a reference.
Unless stated differently, the \gls{RK4} was used as the numerical solver.
For explicit methods, the solver step size $\h$ was set to the sample period $\dt=0.2$.
In \cref{sec:eval-model}, the different model combinations are evaluated on different data sets.
A study on numerical solver implications is presented in  \cref{sec:solver-implication}.

\subsection{Data Sets}
\label{sec:dataset}
For training and testing, two different data sets, \highd\cite{highDdataset} and \round\cite{rounDdataset}, were used.
The data sets contain recorded trajectories from different locations in Germany, including various highways and roundabouts.
The data contain several hours of naturalistic driving data captured at 25 Hz.
Observations for training and inference cover at most 3~s and at least one sample.
During the preprocessing stage, the original input and target data were down-sampled by a factor of 5, effectively setting the sampling period $\dt=0.2$~s.

\subsection{Training and Implementation Details}
All implementations were done using PyTorch \cite{paszke2019pytorch} and PyTorch Geometric \cite{pyg2019}.
For numerical integration, we used \texttt{\small torchdiffeq}~\cite{torchdiffeq}. 
Jacobian calculations were performed using \texttt{\small functorch}~\cite{functorch2021}.
The \emph{Adam} optimizer \cite{kingma2014adam} was employed with a batch size of $128$ and a learning rate of $0.0001$.
A key observation revealed that the number of ground truth states used during training substantially affected test performance. 
For instance, with motion models containing four states, $[x, y, v, \psi]$, despite the dependency of $x$ and $y$ on $v$, having the model concurrently learn to enhance all three as opposed to only $x$ and $y$ led to a notable performance improvement.
It is important to note that the orientation angle cannot be directly incorporated in (\ref{eq:seq_loss}) as it fails to account for the cyclical nature of angles.

\subsection{Evaluation Metrics}
\label{sec:eval_metrics}
Several metrics are used to evaluate the investigated model combinations, which are presented here for a single agent.
These are then averaged over all agents in all traffic situations in the test set.
As the predicted distribution is a mixture, $L^2$-based metrics are computed using the component $j^*$ with the predicted largest weight, $j^* = \argmax_j \mix{j}$.

\begin{itemize}
	\item \emph{\gls{ADE}}:
	\begin{equation}
		\label{eq:ade}
		\text{ADE} = \frac{1}{\predhrz}\sum_{k=1}^{\predhrz} \norm{\stateestim{k} - \state{k}}_2
	\end{equation}
	\item \emph{\gls{FDE}}:
	\begin{equation}
		\label{eq:fde}
		\text{FDE} = \norm{\stateestim{\predhrz} - \state{\predhrz}}_2
	\end{equation}
	\item \emph{\gls{MR}}: The ratio of cases where the predicted final position is not within $2$~m of the ground truth \cite{huang2022survey}.
	\item \emph{\gls{APDE}}:
	\begin{align}
		\label{eq:apde}
		\begin{split}
			\text{APDE} &= \frac{1}{\predhrz}\sum_{k=1}^{\predhrz} \norm{\stateestim{k} - \state{k^*}}_2 \\
			k^* &= \argmin_i \norm{\stateestim{k} - \state{i}}_2
		\end{split}
	\end{align}
	\item \emph{\gls{ANLL}}:
	\begin{equation}
		\text{ANLL} = \frac{1}{\predhrz}\sum_{k=1}^{\predhrz} -\log \left(\sum_j \mix{j}\mathcal{N}(\state{k} | \stateestim{k}^j, \stcov{k}^j)\right)
	\end{equation}
	\item \emph{\gls{FNLL}}:
	\begin{equation}
		\text{FNLL} = -\log \left(\sum_j \mix{j}\mathcal{N}(\state{\predhrz} | \stateestim{\predhrz}^j, \stcov{\predhrz}^j)\right)
	\end{equation}
\end{itemize}

\subsection{Evaluation of Model Combinations}
\label{sec:eval-model}
The prediction model's performance using different motion models on the two data sets is presented in \cref{tab:combined-res}.

\subsubsection{Highway}
\label{sec:results-highd}
There is no significant difference between the employed motion models for the highway trajectory prediction task, with most methods being comparable to the first decimal on $L^2$-based metrics.
However, some minor differences indicate that the \gls{3XI} model, along with all the orientation-based models, performs slightly worse than the others, particularly on NLL.
This is interesting, especially considering that highway trajectories are typically smooth, often described using quintic polynomials in motion planning applications \cite{werling2012optimal}.
By that argument, the \gls{3XI} model should intuitively be competitive to the other methods in this context, which is not the case.
Instead, it is the models \gls{1XI}, \gls{2XI}, \gls{NODE1}, and \gls{NODE2} that report the best results combined over all metrics. 

\begin{table}[!t]
    \caption{Model combination performance by data set}
	\label{tab:combined-res}
	\centering
	\resizebox{\columnwidth}{!}{%
	\begin{tabular}{l l c c c c c c}
		\toprule
		Data set & Model & ADE & FDE & MR & APDE & ANLL & FNLL\\
		\midrule
		 \multirow{11}{*}{\highd{}} & CA & $0.78$ & $2.63$ & $0.55$ & $0.46$ & --- & ---\\ %
		 & CV & $1.49$ & $4.01$ & $0.79$ & $0.87$ & --- & --- \\ \cmidrule(lr{1em}){2-8}
		 & \gls{1XI} & $\bm{0.25}$ & $\bm{0.85}$ & $\bm{0.05}$ & $\bm{0.25}$ & $-1.74$ & $\bm{1.27}$\\ %
		 & \gls{2XI} & $0.28$ & $0.92$ & $0.05$ & $0.28$ & $-1.72$ & $1.57$\\ %
		 & \gls{3XI} & $0.32$ & $1.05$ & $0.06$ & $0.31$ & $4.95$ & $4.30$\\ %
		 & \gls{CL} & $0.32$ & $1.01$ & $0.06$ & $0.31$ & $-0.68$ & $1.99$\\
		 & \gls{C} & $0.33$ & $1.02$ & $0.07$ & $0.31$ & $-0.76$ & $1.97$\\ %
		 & \gls{U} & $0.28$ & $0.92$ & $0.06$ & $0.27$ & $-0.84$ & $1.77$\\ %
		 & \gls{ST} & $0.28$ & $0.93$ & $0.06$ & $0.27$ & $-0.80$ & $1.78$\\ %
		 & \gls{NODE1} & $0.26$ & $0.89$ & $0.06$ & $\bm{0.25}$ & $\bm{-2.03}$ & $1.32$\\ %
		 & \gls{NODE2} & $0.27$ & $0.91$ & $0.06$ & $0.26$ & $-1.90$ & $1.34$\\ %
  		 \midrule
		   \multirow{11}{*}{\round{}} & CA & $4.83$ & $16.2$ & $0.95$ & $3.90$ &  --- & ---\\ %
		   & CV & $6.49$ & $17.1$ & $0.94$ & $4.25$ & --- & ---\\ \cmidrule(lr{1em}){2-8}
		   & \gls{1XI} & $0.99$ & $\bm{3.03}$ & $\bm{0.34}$ & $\bm{0.60}$ & $0.16$ & $\bm{3.31}$\\ %
		   & \gls{2XI} & $0.99$ & $3.10$ & $0.36$ & $0.61$ & $-0.17$ & $3.83$\\ %
		   & \gls{3XI} & $1.45$ & $4.54$ & $0.57$ & $0.96$ & $0.67$ & $4.88$\\ %
		   & \gls{CL} & $1.48$ & $4.81$ & $0.65$ & $1.10$ & $1.22$ & $4.81$\\
		   & \gls{C} & $1.38$ & $4.04$ & $0.74$ & $0.98$ & $0.94$ & $4.40$\\ %
		   & \gls{U} & $1.29$ & $3.85$ & $0.57$ & $0.90$ & $0.93$ & $4.48$\\ %
		   & \gls{ST} & $1.24$ & $3.67$ & $0.56$ & $0.85$ & $0.68$ & $4.01$\\ %
		   & \gls{NODE1} & $0.99$ & $3.07$ & $0.37$ & $0.63$ & $0.20$ & $3.38$\\ %
		   & \gls{NODE2} & $\bm{0.98}$ & $3.09$ & $0.35$ & $0.63$ & $\bm{-0.19}$ & $3.75$\\ %
		\bottomrule
	\end{tabular}}
\end{table}

\begin{figure}[!t]
	\centering
	\subfloat[While the method is capable of multi-agent predictions, the blue car illustrates a single forecast objective.
	The quality of the prediction between motion models varies slightly, although most predictions are close to the ground truth.
	The  \gls{1XI}, \gls{NODE1}, and \gls{NODE2} models are visually the best performing in this scenario.
	The figure background is from the \round{} data set \cite{rounDdataset}.]{\includegraphics[width=\columnwidth]{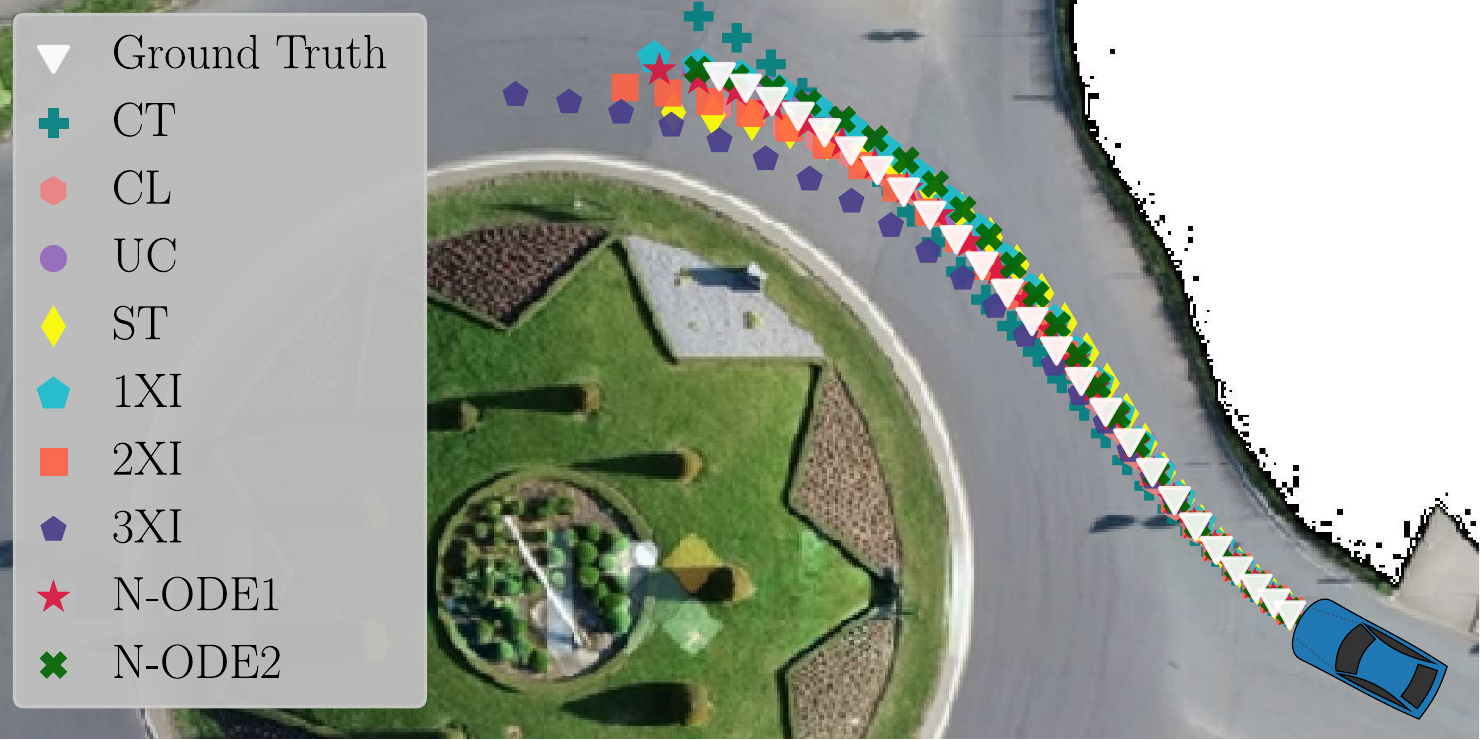}
	\label{fig:round-pred-pos}}
	\hfill
	\subfloat[It is clear from the predictions in (a) that most methods overshoot the true trajectory.
	Comparing how well the model predicts the velocity may provide an indication of which models are the most effective.]{\includegraphics[width=0.95\columnwidth]{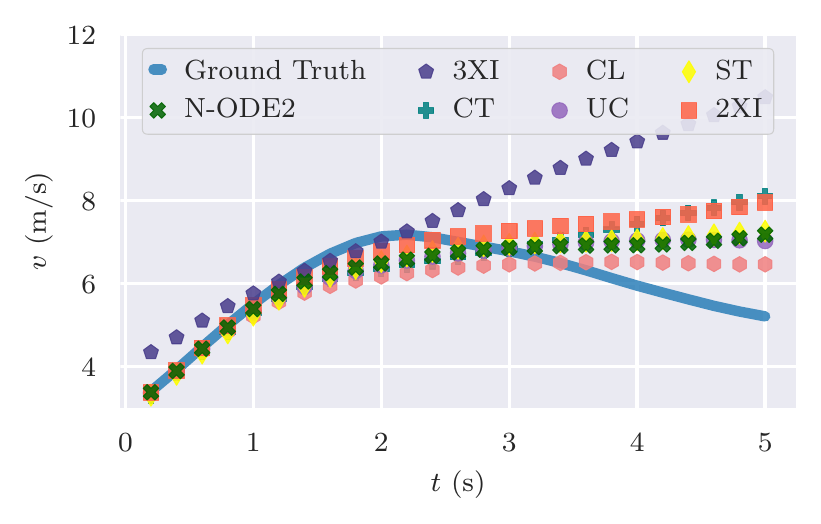}
	\label{fig:round-pred-vel}}
	\caption{Example predictions of the motion models on a roundabout scenario.}
	\label{fig:round-pred}
    \vspace{-0.05in}
\end{figure}

\subsubsection{Roundabout}
The performance of the methods when predicting roundabout trajectories, compared to the results for highway predictions, indicates that this is a more complex problem.
Regardless, most of the deep-learning enhanced motion models perform well.
The orientation-based models are underperforming in this context, reporting errors significantly larger than the others.
This is interesting, given that these types of motion models are typically used in related works \cite{cui2020deep,salzmann2020trajectron, li2021spatio}.
Instead, the combined results indicate that simple integrators are adequate for accurate learning-based motion prediction, which is encouraging given their reduced complexity and low computational demands.
The prediction performance can be further assessed by studying the illustration in \cref{fig:round-pred}, showing an example prediction using test data from the \round{} data set.
In particular, \cref{fig:round-pred-vel} illustrates the difficulties of the investigated problem, but also the limitations of the methods.
Over a $5$~s prediction horizon, numerous events can occur, and although the methods accurately capture the overall true path, only a few successfully predict deceleration entering the curve.
Still, most models remain accurate for up to $3$~s before diverging.

\subsection{Implications of Numerical Solver Selection}
\label{sec:solver-implication}
Since the choice of integration method could substantially impact the solution trajectory, it is interesting to study its effect within this context.
Three different motion models were selected for this study: \gls{ST}, \gls{2XI}, and \gls{NODE2}.
The models were trained and evaluated using different numerical solvers: \gls{EF}, \gls{Heun}, \gls{RK3}, \acrlong{RK4} (\gls{RK4}), \gls{RK45}, and an \gls{IMA} \cite{ascher1998computer}.
Both data sets were subjected to the investigation but yielded similar results.
While the conclusions hold for both scenarios, only the performance on \round{} will be discussed here.

\begin{table}[!t]
	\caption{RounD solver performance}
	\label{tab:round-solver-imp}
	\centering
	\resizebox{\columnwidth}{!}{%
	\begin{tabular}{l l c c c c c c}
		\toprule
		Model & Solver & ADE & FDE & MR & APDE & ANLL & FNLL\\
		\midrule
		 \multirow{6}{*}{\gls{ST}} & EF & $1.82$ & $4.97$ & $0.73$ & $1.24$ & $2.32$ & $4.71$\\ %
				& Heun & $\bm{1.23}$ & $\bm{3.64}$ & $\bm{0.55}$ & $\bm{0.83}$ & $0.74$ & $4.10$\\ %
				& RK3 & $1.28$ & $3.80$ & $0.58$ & $0.87$ & $0.75$ & $4.12$\\ %
				& RK4 & $1.24$ & $3.67$ & $0.56$ & $0.85$ & $\bm{0.68}$ & $\bm{4.01}$\\ %
				& \gls{RK45} & $1.30$ & $3.84$ & $0.67$ & $0.90$ & $0.91$ & $4.42$\\ %
				& Adams & $1.24$ & $3.69$ & $0.56$ & $0.84$ & $0.73$ & $4.12$\\ %
  		 \midrule
		   \multirow{6}{*}{\gls{2XI}} & EF & $1.50$ & $4.18$ & $0.58$ & $0.89$ & $5.67$ & $7.10$\\ %
		   & Heun & $\bm{0.97}$ & $\bm{3.07}$ & $\bm{0.34}$ &  $\bm{0.60}$ & $\bm{-0.22}$ & $3.75$\\ %
		   & RK3 & $0.98$ & $3.08$ & $0.35$ & $0.61$ & $\bm{-0.22}$ & $3.77$\\ %
		   & RK4 & $0.99$ & $3.10$ & $0.36$ & $0.61$ & $-0.17$ & $3.83$\\ %
		   & \gls{RK45} & $0.99$ & $3.09$ & $0.36$ & $\bm{0.60}$ & $-0.19$ & $3.77$\\ %
		   & Adams & $1.01$ & $3.12$ & $0.39$ & $0.61$ & $-0.17$ & $\bm{3.73}$\\ %
			\midrule
		   \multirow{6}{*}{\gls{NODE2}} & EF & $1.90$ & $7.48$ & $0.72$ & $1.07$ & $5.84$ & $7.48$\\ %
				& Heun & $1.00$ & $3.11$ & $\bm{0.35}$ & $0.62$ & $\bm{-0.19}$ & $3.76$\\ %
				& RK3 & $0.99$ & $3.10$ & $0.37$ & $0.62$ & $\bm{-0.19}$ & $3.77$\\ %
				& RK4 & $\bm{0.98}$ & $3.09$ & $\bm{0.35}$ & $0.63$ & $\bm{-0.19}$ & $3.75$\\ %
				& \gls{RK45} & $1.10$ & $3.46$ & $0.44$ & $0.70$ & $0.24$ & $4.08$ \\
				& Adams & $\bm{0.98}$ & $\bm{3.04}$ & $\bm{0.35}$ & $\bm{0.61}$ & $\bm{-0.19}$ & $\bm{3.73}$\\ %
		\bottomrule
	\end{tabular}}
\end{table}

The effect of numerical solver selection is presented in \cref{tab:round-solver-imp}.
The results illustrate varying model responses to different solvers.
What is true for all motion models, however, is that the worst results are achieved in combination with \gls{EF}.
Interestingly, using a second-order solver, like Heun's, has a significant positive impact on performance compared to \gls{EF}.
Going beyond the second-order method, offers only marginal performance increases, much of which is dependent on the model--solver combination.
For example, \gls{NODE2} responded positively toward using the implicit Adam's method, but arguably not enough to be considered significant.

\begin{figure}[!t]
	\centering
	\includegraphics[width=\columnwidth]{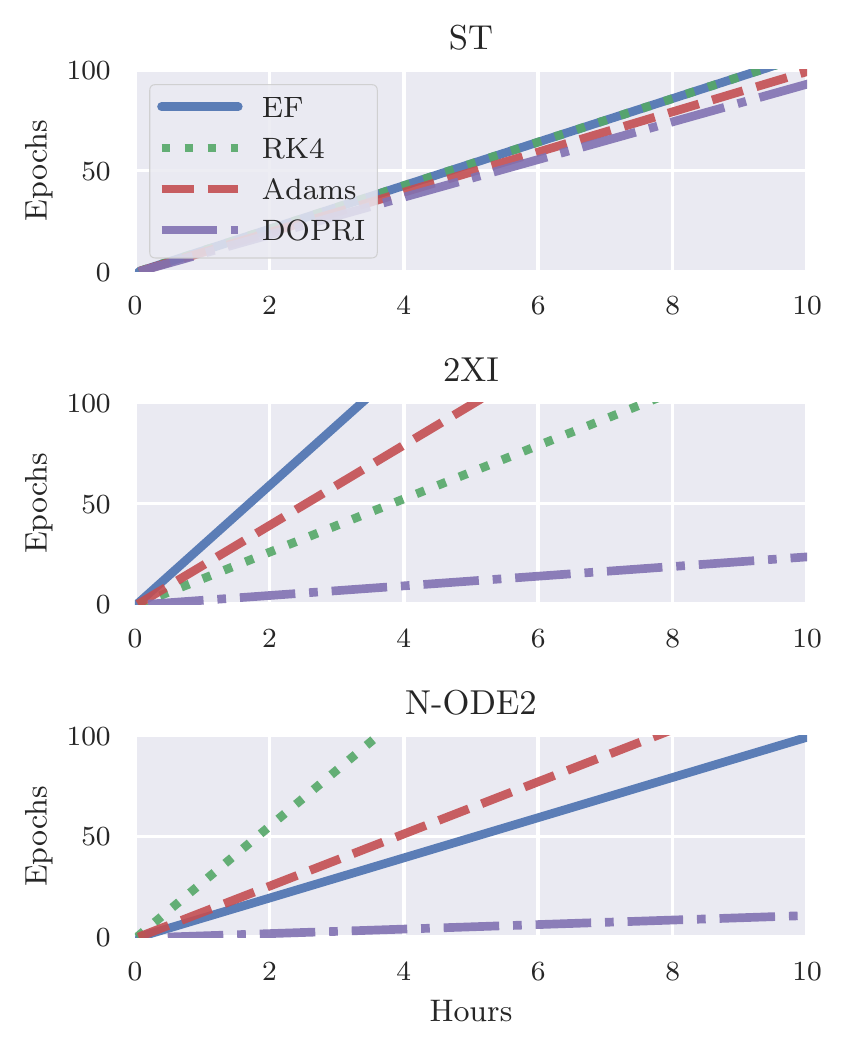}
	\caption{Training progress of \gls{ST}, \gls{2XI}, and \gls{NODE2} over 100 epochs for different solvers on the \round{} data set, using one NVIDIA A100 GPU. 
	The training process for the various models is differently affected by solver selection, especially when using the variable-step solver, \gls{RK45}.
    Interestingly, training \gls{NODE2} with \gls{EF} resulted in a longer total duration than using \gls{RK4}. 
    This is possibly because of the combination of the small stability region of \glspl{EF} and the \glspl{NODE2} dynamics.
    Instead the additional function evaluations \gls{RK4} adds increased stability during training.
	Although excluded from the figure for clarity, solvers like Heun's and RK3 were in magnitudes similar to EF.}
	\label{fig:solver_training}
\end{figure}

The trade-off between performance and computational demands is an interesting topic.
In \cref{fig:solver_training}, the training performance over $10$ hours using the aforementioned models and selected solvers is presented.
The choice of numerical solver can have a significant impact on the training duration, likely resulting from additional function evaluations specific to higher-order methods.
As one might expect, employing the variable-step solver \gls{RK45} led to a notably extended training duration.

\begin{figure}[!t]
	\centering
	\includegraphics[width=\columnwidth]{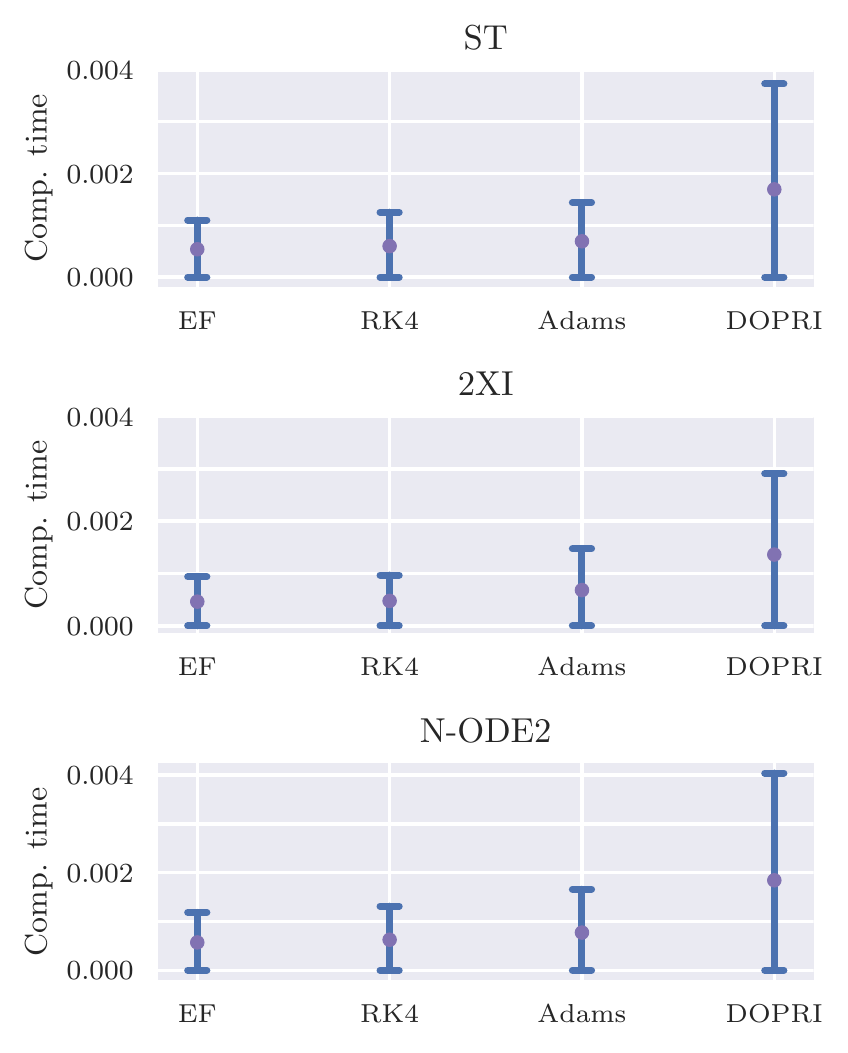}
	\caption{Inference time per forecast objective using \gls{ST}, \gls{2XI}, and \gls{NODE2} with different solvers on the \round{} test set, using a single NVIDIA A100 GPU. 
	The data points (graph sequences) consist of a varying number of agents, leading to variability in computation time. To represent this variation, a $95\%$ confidence interval is employed.}
	\label{fig:inference_time}
\end{figure}

Although the effect on model inference time is not as pronounced as in the case of training performance, similar trends can be observed (see \cref{fig:inference_time}).
Most model-solver combinations require less than $1$ millisecond to compute predictions for a single agent, with little variation.
However, as with the training duration, using the variable-step solver \gls{RK45} necessitated a much longer computation time.

		\section{Conclusion}
An evaluation of differentially-constrained motion models and numerical solvers for learning-based trajectory prediction was presented.
Using the MTP-GO framework, a range of motion models, from pure integrators to kinematic models and neural ODEs, were examined.
It was found that simpler models, such as low-order integrators, yielded the best results irrespective of scenario complexity.
The study also revealed that the selection of a numerical solver can significantly influence both training and prediction performance.
Although the findings indicate that effective prediction performance can be achieved using a second-order numerical solver like Heun's method, it underscores the importance of making well-informed model and solver choices when implementing these methods.
		\section*{Acknowledgments}
Computations were enabled by the supercomputing resource Berzelius provided by National Supercomputer Centre at Linköping University and the Knut and Alice Wallenberg foundation.

		\bibliographystyle{IEEEtran}
		\bibliography{IEEEabrv,references.bib}{}

	\end{document}